\documentclass[twoside]{article}

\usepackage{amssymb}
\usepackage{amsmath}
\usepackage{dsfont}
\usepackage{svg}
\usepackage{amsthm}   
\usepackage{graphicx}
\usepackage{subcaption} 
\usepackage{bm}
\usepackage{booktabs}
\usepackage{tabularx}

\usepackage{algorithm}
\usepackage{algorithmicx}
\usepackage{algpseudocode}
\usepackage{listings}
\usepackage{placeins}
\usepackage{hyperref}

\usepackage[nolist,nohyperlinks]{acronym}
\begin{acronym}
\acro{dnn}[DNN]{deep neural network}
\acro{mcmc}[MCMC]{Markov Chain Monte Carlo}
\acro{ssr}[SSR]{semi-structured regression}
\acro{hmc}[HMC]{Hamilton Monte Carlo}
\end{acronym}

\newcommand{\vx}{\mathbf{x}} 
\newcommand{\vu}{\mathbf{u}} 
\newcommand{\vw}{\mathbf{w}}
\newcommand{\vp}{\mathbf{p}}

\newcommand{\bl}{\mathbf{b}_{\boldsymbol{\Lambda}}}
\newcommand{\vt}{\boldsymbol{\theta}}
\newcommand{\vphi}{\boldsymbol{\varphi}}
\DeclareMathOperator{\Pois}{Pois}

\usepackage[accepted]{aistats2024}
\usepackage[round]{natbib}

\begin{document}

%

%

\twocolumn[

\aistatstitle{Bayesian Semi-structured Subspace Inference}

\aistatsauthor{ Daniel Dold \And David Rügamer \And Beate Sick \And Oliver Dürr}

\aistatsaddress{ HTWG Konstanz \And LMU Munich and MCML \And UZH and ZHAW Zürich \And  HTWG Konstanz} ]

\begin{abstract}
    Semi-structured regression models enable the joint modeling of interpretable structured and complex unstructured feature effects. The structured model part is inspired by statistical models and can be used to infer the input-output relationship for features of particular importance. The complex unstructured part defines an arbitrary deep neural network and thereby provides enough flexibility to achieve competitive prediction performance. While these models can also account for aleatoric uncertainty, there is still a lack of work on accounting for epistemic uncertainty. In this paper, we address this problem by presenting a Bayesian approximation for semi-structured regression models using subspace inference. To this end, we extend subspace inference for joint posterior sampling from a full parameter space for structured effects and a subspace for unstructured effects. Apart from this hybrid sampling scheme, our method allows for tunable complexity of the subspace and can capture multiple minima in the loss landscape. Numerical experiments validate our approach's efficacy in recovering structured effect parameter posteriors in semi-structured models and approaching the full-space posterior distribution of MCMC for increasing subspace dimension. Further, our approach exhibits competitive predictive performance across simulated and real-world datasets.
\end{abstract}

\section{INTRODUCTION}
\begin{figure*}[!t]
    \centering
    \includegraphics[width=\textwidth]{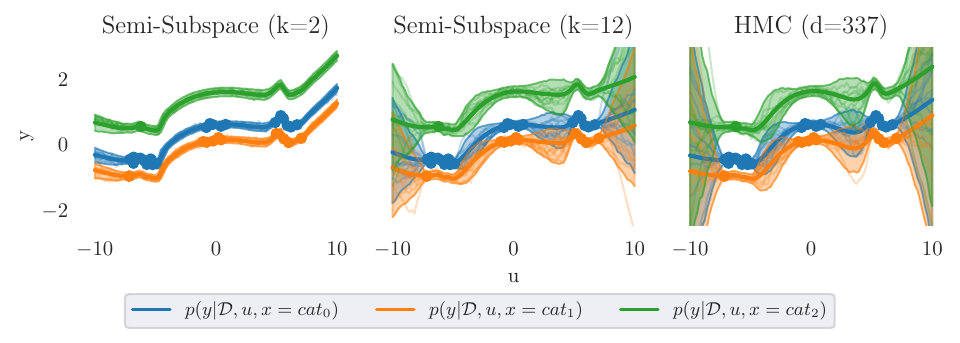}
    \caption{Comparison of semi-structured subspace inference and \ac{hmc} for an SSR model. The SSR is defined as a combination of a linear shift induced by the categorical feature $x$ (color code) and a non-linear trend in $u$ (x-axis) modeled by a deep neural network (cf.~Equation \ref{alg:1}). 
    Left/center: posterior predictive for dataset $\mathcal{D}$ and outcome $y$ with a 2-dim.~and 12-dim.~subspace; right: posterior predictive of \ac{hmc} without any approximation. Points represent the data, colored by their category of $x$, the solid line is the mean, and shading depicts the 95\% Highest Density Interval.}
    \label{fig:fig1}
\end{figure*}

A linear model is inherently transparent and interpretable due to its model structure and underlying assumptions. When given features denoted as $\vx$, the expected outcome $\mathbb{E}(y|\vx)$ for a variable of interest $y\in\mathbb{R}$ is estimated as a linear combination $\vx^\top \vt$ of features $\vx \in \mathbb{R}^p$ and corresponding parameters $\vt\in\mathbb{R}^p$. 
This interpretability is also preserved in the various extensions such as generalized linear models \citep[GLMs;][]{nelder1972generalized} for non-Gaussian conditional outcome distributions or generalized additive models \citep[GAMs;][]{wahba1990spline,wood2017generalized} for the inclusion of non-linearity via splines. 
While these extensions allow for a flexible definition of univariate or moderate-dimensional multivariate feature effects, they lack flexibility for complex higher-order interactions and are restricted to tabular features.
A \ac{dnn}, on the other hand, learns complex feature effects in a data-driven fashion and can work for different input modalities (e.g., image data). 
The fusion of a structured and interpretable statistical model with highly flexible \acp{dnn} thus has some attractive properties and has been investigated over the last 30 years (cf.~Section~\ref{sec:ssr}).

While this combination is flexible and attractive from a modeling point of view, many properties of this so-called \ac{ssr} are yet still unexplored. One important aspect is their uncertainty quantification, particularly relevant in their application in the medical domain \citep{dorigatti_frequentist_2023}. Although some of the more recent approaches account for aleatoric \citep{rugamer_semi-structured_2022} or epistemic uncertainty in \ac{ssr} models \citep{durr_bernstein_2022,dorigatti_frequentist_2023}, all existing approaches do either not account for the epistemic uncertainty arising from the model's \ac{dnn} part or assume this uncertainty to be given.
Another significant but understudied challenge in traditional \ac{ssr} models is the joint optimization of the two model components.
On the one hand, \acp{dnn} can theoretically fit the training data perfectly, potentially leaving little to explain for the structured part\citep{zhang2017understanding, zhang2021understanding}. Optimization of structured models such as GLMs, on the other hand, is typically done using more advanced second-order methods. This optimization asymmetry in \ac{ssr} complicates the process of joint optimization.

The Bayesian paradigm offers a rigorous framework for quantifying uncertainty and \ac{mcmc} methods, relying on sampling rather than optimization, are often considered as the gold standard for inference in Bayesian neural networks \citep{wiese2022}. Hence, a Bayesian variant of \ac{ssr} models could provide inference statements and circumvent the aforementioned issues with the joint optimization of structured and \ac{dnn} model parts. These sampling-based approaches, however, are computationally intensive and struggle with high-dimensional parameter spaces typical for \ac{dnn}s. 

\textbf{Our Contribution} In this work, we present \emph{semi-structured subspace inference}, a sampling-based method that not only captures aleatoric and epistemic uncertainty in \ac{ssr} models but also addresses the optimization asymmetry often observed in such models. Our method allows obtaining the posterior for every structured model parameter while accounting for the \ac{dnn}'s uncertainty. By using an adjustable subspace approximation of the \ac{dnn} part, it is compatible with common \ac{mcmc} methods. We show that semi-structured subspace inference 1) yields nearly the same posterior distribution as full-space \ac{mcmc} methods for the structured model component, and 2) provides posterior predictive distributions of the quality of full-space inference even when using a highly-compressed subspace (see Figure~\ref{fig:fig1}). 
We further provide numerical evidence confirming the efficacy of our approach and superiority when compared to other Bayesian approximation methods. 

\section{RELATED WORK}
Before introducing our method in Section~\ref{sec:method}, we briefly introduce \ac{ssr} and Subspace inference in the following.

\subsection{Semi-Structured Regression} \label{sec:ssr}
The fusion of structured models from statistics and (deep) neural networks started with \citet{ciampi1995designing,ciampi1997statistical}, followed by extensions to model generalized additive neural networks \citep{potts1999generalized,de2007generalized,De.2011}.
In recent years, this combination has returned to the limelight under the name of \emph{wide and deep learning} \citep{Cheng.2016} or semi-structured regression \citep[SSR;][]{pho}. Due to its flexibility, SSR has been adapted for various scenarios such as Deep GLMs \citep{Tran.2018}, Deep Bayesian regression \citep{Hubin.2018}, survival analysis  \citep{Poelsterl.2020,kopper2021}, state space models \citep{amoura2011state}, transformation models \citep{baumann.2020, sick2021deep}, ordinal  \citep{KOOK2022108263} or distributional regression \citep{rugamer_semi-structured_2022}. The question of how uncertainty can be quantified in a combination of an (unstructured) DNN and a structured regression model has however received not much attention. Only recently, \citet{dorigatti_frequentist_2023} showed that for given DNN uncertainty, it is possible to derive the uncertainty for the structured model parameters in SSR models in a frequentistic manner. While their derived confidence intervals achieve nominal coverage when the deep uncertainty quantification method works well, they also point out the failure of the method if the uncertainty of the DNN is not well quantified. Their approach further leaves various points unanswered as it only focuses on the structured parameter uncertainty and cannot be embedded in a Bayesian setting despite many of the DNN uncertainty quantification methods being motivated in a Bayesian context \citep[e.g.,][]{daxberger2021laplace,izmailov_subspace_2020-1}. Another option to account for uncertainties and improve model performance are deep ensembles \citep{lakshminarayanan2017simple}. While deep ensembling was adapted for semi-structured models \citep{kook_deep_2022}, it only accounts for algorithmic uncertainty in the \ac{dnn} model part and cannot be considered fully Bayesian.
\subsection{Bayesian Approximations and Subspace Inference}
In complex Bayesian models that have many parameters, and where \ac{mcmc} is not computationally feasible, Laplace approximation \citep{daxberger2021laplace} provides a tractable alternative. This method approximates the posterior distribution with a Gaussian distribution centered at a single mode. However, this simplification neglects the potentially multi-modal nature of the posterior in complex models. 
In contrast, subspace inference \citep{izmailov_subspace_2020-1} provides an approach capable of capturing multiple modes in the posterior. This is achieved by defining a lower-dimensional subspace within the parameter space that can accommodate multiple modes. This subspace facilitates efficient posterior sampling using \ac{mcmc} methods. Two methods were proposed to construct this subspace within the high-dimensional weight space of a neural network: the first employs principal component analysis on weights collected during the last training epochs, which typically corresponds to a single minimum in the loss space and hence captures a single mode of the posterior; the second method from \citet{garipov_loss_2018} connects two local minima in the loss landscape using a quadratic Bézier curve, enriching the model's uncertainty by potentially capturing multiple posterior modes. 
Control points of the Bézier curve are determined by weights resulting from two independent training runs, with a third optimization refining the last control point to ensure all weights along the curve yield well-performing models. 
\citet{izmailov_subspace_2020-1} empirically showed that using a quadratic Bézier with two connected modes outperformed the principal component analysis approach. Thus our work exclusively adopts the Bézier approach for subspace inference. While effective, this method restricts the subspace dimension to two dimensions and necessitates multiple training runs. \citet{wortsman2021learning} improved upon this by developing an algorithm that works within a single training run, but also uses the quadratic Bézier curves. 
To the best of our knowledge, there has been no further development building on \citet{wortsman2021learning} or extending the Bézier curve approach originally introduced by \citet{garipov_loss_2018}. Recent work by \citet{jantre2023learning} incorporates output information but remains confined to exploring a single mode of the posterior.

\section{SEMI-STRUCTURED SUBSPACE INFERENCE} \label{sec:method}

A semi-structured regression model is defined as an additive combination of a structured model part, capturing the interpretable effect of tabular input features $\vx\in\mathbb{R}^p$, and an unstructured model part processing complex effects of a potentially complex input $\mathbf{u}\in\mathcal{U}$ through a DNN. While the class of models is not restricted to certain regression models and the structured model part can be flexibly defined, we here focus on mean regression approaches, where the mean $\mu$ of some distribution is modeled as a semi-structured predictor of the form
\begin{equation} \label{eq:ssr}
    \mu = \vx^\top \vt + \text{DNN}(\mathbf{u}).
\end{equation}
In \eqref{eq:ssr}, $\vt\in\mathbb{R}^p$ are the interpretable parameters of the structured model part and $\text{DNN}: \mathcal{U} \to \mathbb{R}$ is parametrized with weights $\vw\in\mathbb{R}^d$, where $d$ is usually large. We use this simple definition for better readability but note that extending our approach to models with more complex structured effects such as splines or distribution regression approaches as discussed in Section~\ref{sec:ssr} is straightforward. As the parameters $\vt$ of the structured part play a special role, quantifying their uncertainty is one of our primary goals. One option to quantify the uncertainty is to employ a Bayesian approach.

\paragraph{Na\"ive Approximation Methods}
Translating SSR models into a Bayesian framework requires some form of approximation as classical \ac{mcmc} is infeasible for the model's unstructured \ac{dnn} part. A na\"ive approach to implement approximation techniques such as Laplace approximation or subspace inference for \ac{ssr} models would be to treat the parameters from the structured model part $\vt$ without special attention.
This would mean simply extending the $d$-dimensional \ac{dnn} weight space with the $p$-dimensional space of $\vt$ and applying the original approximation to the combined $(d+p)$-dimensional space without further modification. 
However, by not differentiating between the small parameter set $\vt$ and the overparameterized weights of a DNN, these naive approaches constrain the posterior $p(\vt|\mathcal{D})$ in its shape (e.g., a unimodal Gaussian distribution for Laplace approximation) or flexibility (cf.~Figure~\ref{fig:naive_post_compare} for subspace inference).

\paragraph{Semi-Structured Subspace Inference}
Our method is an extension of subspace inference, originally introduced by \citet{izmailov_subspace_2020-1}, tailored specifically for \ac{ssr} models. 
The core premise of our approach is a dimensionality reduction applied to the weight vector \( \vw \) of the neural network to a $k$-dimensional subspace, allowing sampling in \( \mathbb{R}^k \times \mathbb{R}^p \), with \( k \ll d \), instead of \( \mathbb{R}^d \times \mathbb{R}^p \).
In the following, we elaborate on how our method incorporates elements from \cite{izmailov_subspace_2020-1} while introducing our own adaptations and enhancements to better suit \ac{ssr} models.
The design of the sampling space is guided by the following criteria. First, the subspace must be sample-efficient, concentrating on regions of the loss landscape with low loss values. Second, it must encompass a diverse set of weight configurations that correspond to small loss values. Third, the subspace should facilitate smooth and rapid traversal between distinct low-loss regions, thereby enabling the exploration of a diverse set of solutions. 
Finally, the subspace construction must be aware of the structured model part \( \vx^\top \vt \) as \(  \vx^\top \vt \) can have a significant impact on the loss landscape of the DNN component. 

%
\subsection{Construction of the Approximate Sampling Space}
\label{sec:method_subspace_construct}
\begin{figure}[h]
    \centering
    \includegraphics[width=\linewidth]{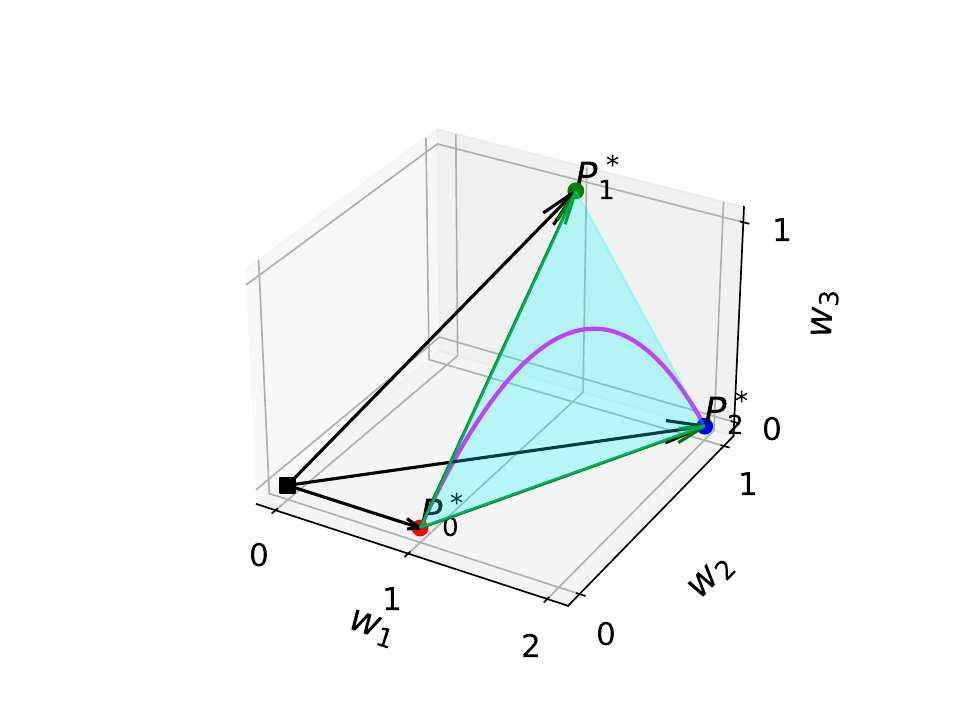}  
    \caption{Bézier curve (magenta) in three-dimensional weight space, controlled by optimized points $\vp_0^*, \vp_1^*, \vp_2^*$, which form a two-dimensional subspace $\text{AffSpan}(\vp_0^*, \vp_1^*, \vp_2^*)$ indicated by the cyan triangle that includes the Bézier curve. The difference vectors $\vp_1^* - \vp_0^*$ and $\vp_2^* - \vp_0^*$ spanning the affine subspace are shown in green.}
    \label{fig:sub_space}
\end{figure}

In line with these guiding principles, we introduce a parametric path \( \bl: [0, 1] \rightarrow \mathbb{R}^d \) that interconnects weight vectors $\vp_l$, $l={0\ldots k}$, of $k+1$ neural network parametrizations, within the $d$-dimensional DNN weight space (see Figure \ref{fig:sub_space}). This path is formulated using a Bézier curve with the weight vectors \(\vp_l\) serving as control points:
\begin{equation}
    \bl(t) = \sum_{l=0}^{k} \binom{k}{l} (1-t)^{k-l} t^l \vp_l.
    \label{eq:bez}
\end{equation}
The parameterization of the curve is given by $\boldsymbol{\Lambda}:=(\vp_0,\ldots, \vp_k)$. 
Each point $\bl(t)$ together with the parameters $\vt$ comprises a parameter set of an SSR model with loss ${\cal L}(\bl(t), \vt)$. 
Empirical evidence, as outlined in \cite{garipov_loss_2018}, indicates the presence of a low-loss valley between distinct SGD solutions. Following their approach, we minimize the functional \( L(\vt,\boldsymbol{\Lambda}) \) defined as
\begin{equation}
L(\vt,\boldsymbol{\Lambda}) = \int_0^1 \mathcal{L}(\bl(t), \vt) \, dt.
\label{eq:objective}
\end{equation}
We compute an unbiased estimate of the objective \eqref{eq:objective} by sampling \( t \sim U(0,1) \) and update \( (\vt,\boldsymbol{\Lambda}) \) via mini-batch gradient descent (Algorithm \ref{alg:1}).
\begin{algorithm}
\caption{Subspace construction}
\label{alg:1}
\begin{algorithmic}[1]
\State \textbf{Initialize} weights $\vp_0, \ldots, \vp_k$ and \(\vt\) randomly
\While{validation loss still reducible}
    \For{each minibatch $\mathcal{B}$ of training data $\mathcal{D}$}
    \State Sample \(t \sim U(0, 1)\)
    \State Compute \(\mathcal{L}(\bl(t), \vt)\) and gradients  \(\nabla \mathcal{L} \)
    \State {Update \(\vp_0, \ldots, \vp_k\) and \(\vt\) using any  
    \Statex \hspace{\algorithmicindent}  \hspace{\algorithmicindent}\hspace{\algorithmicindent}SGD variant (e.g., Adam)}
    \EndFor
\EndWhile
\State \textbf{return} optimized \(\vp_0^*, \ldots, \vp_k^*\) and \(\vt^*\)
\end{algorithmic}
\end{algorithm}
The optimal parameters $\boldsymbol{\Lambda}^*$ define a \(k\)-dimensional (affine) subspace in \(\mathbb{R}^d\)
\begin{equation}
\text{AffSpan}(\boldsymbol{\Lambda}^*) = \left\{ \vp_0^* + \sum_{i=1}^{k} \varphi'_i \Delta \vp^*_i \,\Big| \, \varphi'_i \in \mathbb{R} \right\} \;,
\label{eq:aff_span}
\end{equation}    
where $\Delta \vp^*_i = (\vp_i^* - \vp_0^*)$, and $\boldsymbol{\Lambda}^* =(\vp_0^*, \ldots, \vp_k^*)$, which includes the estimated low-loss valley given by the Bézier curve (see Supplementary Material~\ref{proof:bezier} for a proof).
Figure \ref{fig:sub_space} illustrates this concept for \(k=2\) and \(d=3\).
In contrast to \citet{izmailov_subspace_2020-1}, our approach streamlines subspace construction by training the Bézier curve model in a single stage, eliminating the need for sequential training of $\vp^*_0$, $\vp^*_2$, and $\vp^*_1$ with fixed $\vp^*_0$ and $\vp^*_2$.


\subsection{Subspace Sampling}
In order to approximate the DNN part's posterior, we sample weights in the $k$-dimensional affine subspace defined in \eqref{eq:aff_span}.
%
%
%
To ensure compatibility with the methodology presented in \cite{izmailov_subspace_2020-1} for the case \(k=2\), we adopt an orthogonal coordinate system for sampling, as opposed to directly sampling \(\varphi'\) in \eqref{eq:aff_span}. 
Specifically, we perform a translation to center the controlling points \(\mathbf{p}^*_i\) around their mean vector, given by 
$
\bar{\mathbf{p}}= \frac{1}{k+1} \sum_{i=0}^{k} \mathbf{p}_i^*.
$
Subsequently, we perform a principal component analysis on the centered points to construct an orthogonal projection matrix, denoted as \(\boldsymbol{\Pi}: \mathbb{R}^k \rightarrow \mathbb{R}^d\). 
This matrix \(\boldsymbol{\Pi} \in \mathbb{R}^{d \times k}\) encapsulates the first \(k\) principal components of the centered dataset.
Given a sample vector \(\boldsymbol{\varphi} = (\varphi_1,\ldots,\varphi_k) \in \mathbb{R}^k\), we can then transform $\boldsymbol{\varphi}$ into a weight vector $\mathbf{w}$ in the \(d\)-dimensional weight space of the neural network via:
\begin{equation}
    \mathbf{w} = \bar{\mathbf{p}} + \boldsymbol{\Pi} \boldsymbol{\varphi}.
    \label{eq:sampling_projection}
\end{equation}
%
Together with the structured parameters $\vt$, the sampling procedure hence involves generating samples from a tuple \((\boldsymbol{\varphi}, \vt)\). 
This allows us to compute the likelihood contribution of the DNN part and the structured part as:
%
\begin{equation}
    p(\mathcal{D}|\vt, \boldsymbol{\varphi}) = p(\mathcal{D}|\vt, \vw = \bar{\vp} + \boldsymbol{\Pi} \boldsymbol{\varphi}).
    \label{eq:likelihood_semi-subspace}
\end{equation}
Since both $p$ and $k$ are reasonably small, our method highly supports efficient sampling with \ac{mcmc} algorithms including sophisticated algorithms such as \ac{hmc}. 
The sampling space design inherently captures multiple low-loss regions, fostering effective sampling and diverse solution exploration.
Within this space, through the Bézier curve, a low-loss pathway is embedded to ensure smooth transitions during sampling. Importantly, this design integrates the interpretable parameter $\vt$ into the sampling space without being directly affected by the approximation.
%
%
\paragraph{Priors}
As in \cite{izmailov_subspace_2020-1}, we model the vectors \( \boldsymbol{\varphi}\) and \(\vt\) using independent multivariate normal distributions \(\boldsymbol{\varphi} \sim \mathcal{N}(0, \boldsymbol{I}_k \sigma_{\varphi}) \) and \(\vt \sim \mathcal{N}(0, \boldsymbol{I}_p \sigma_\theta) \). Although a Gaussian prior was found to be adequate for \(\vt\) in our studies, our framework accommodates the use of more complex priors. This flexibility is particularly beneficial for interpreting the structured model component parameterized by \(\vt\). To bring the lower-dimensional subspace priors in line with conventional Bayesian \ac{dnn} priors, \cite{izmailov_subspace_2020-1} discussed the use of temperature scaling. A detailed discussion for \ac{ssr} models is provided in the Supplementary Material \ref{app:temp}.

\section{NUMERICAL EXPERIMENTS}
We now illustrate the advantages of our framework on four experiments where we compare our approach with 1) a na\"ive Bayesian SSR approximation on a simple regression toy experiment; 2) ground truth results derived from \ac{hmc} on simulated data; 3) MCMC and approximation methods on benchmark datasets, and 4) various \ac{ssr} approaches on a complex medical dataset.
For the latter, we employ the Elliptic Slice Sampler~\citep{murray2010elliptical}. 
For all other cases, full batch processing is possible and we hence choose \ac{hmc} to sample in the subspace as it typically results in a larger effective sample size. Further details and experimental results can be found in the Supplementary Material \ref{sec:app:add_results}.

\subsection{Comparison with Na\"ive Subspace Inference}
\label{sec:exp_naive_sub}
In this first experiment, we aim to compare the posterior distributions derived from the structured model component of our approach against those from a na\"ive subspace approximation. To this end, we adapt the synthetic dataset \( \{f(u_i), u_i\}_{i=1}^{n_{\text{train}}} \) from \cite{izmailov_subspace_2020-1} with noisy nonlinear function \( f \). For each data point $i$, we then incorporate a structured effect by randomly choosing a category vector $\vx_i \in \{(0,0)^\top, (1,0)^\top, (0,1)^\top\}$ to shift $f(u_i)$ by an offset $\vx_i^\top \vt^*$ with $\vt^* = (-0.5, 1)^\top$, resulting in the final training dataset \( \{ y_i = f(u_i) + \vx_i^\top \vt^*, u_i, \vx_i\}_{i=1}^{n_{\text{train}=35}} \).

Knowing the structure of the data-generating process, we model the simulated data by a corresponding \ac{ssr} as in Equation \eqref{eq:ssr}. 
For $\text{DNN}(u)$ we choose a simple network with two fully-connected layers, each with 16 neurons and ReLU activation, and a linear output layer.
Inference is done by sampling weights from a $k$-dimensional subspace for the unstructured model part while sampling from the full space of $\vt = (\theta_0,\theta_1)^\top$ for the structured model part.
The na\"ive approach does not differentiate between the two model parts and applies subspace inference on the combined $(w,\theta)$-space.

\begin{figure}[htbp]
    \centering
    \includegraphics[width=\linewidth]{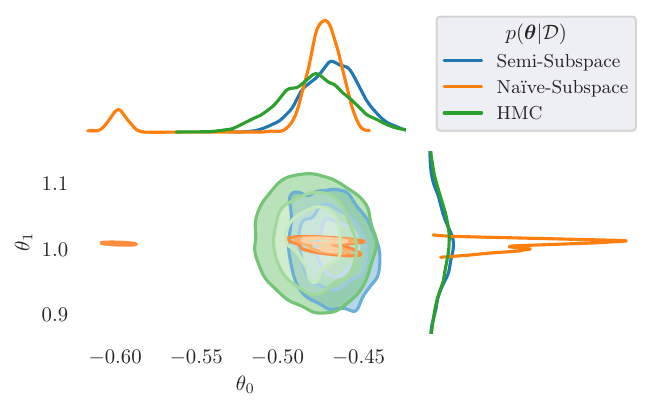} 
    \caption{Posterior of the parameters in the structured model part using the na\"ive subspace approximation with $k=4$ (Na\"ive-Subspace), our approach with $k=2$ and $p=2$ (Semi-Subspace), and \ac{hmc} running in the full parameter space. The top and the right plot shows the marginal posterior distribution, whereas the center plot visualizes the bivariate distribution using a kernel density estimator based on 4000 samples from 10 \ac{hmc} chains.}
\label{fig:naive_post_compare}
\end{figure}

\textbf{Results}\quad As shown in Figure~\ref{fig:naive_post_compare}, the na\"ive subspace approach fails to represent the true posterior accurately, this is especially visible along the $\theta_1$ direction. 
We further visualize the resulting posterior predictive in Figure~\ref{fig:fig1}. While the choice of $k$ does not influence the predictive distribution's mean, we find that $k=2$ produces a too narrow distribution compared to the gold standard obtained via full space \ac{hmc}, and hence an overconfident uncertainty measure for both in- and out-of-distribution data. Notably, when increasing to $k=12$, which is still much less than $d=337$, we obtain almost the same level of uncertainty as \ac{hmc} despite reducing the space by a factor of around 28. In contrast, the Laplace approximation underestimates the epistemic uncertainty (cf.~Supplementary Material, Section~\ref{sec:app:add_results_toydata}). This is also reflected in the log pointwise predictive density (LPPD)\footnote{To be comparable with \cite{wiese2022}, we divided each reported LPPD by the number of data points} evaluated on $n_\text{test} = 365$ test data points, where the Laplace approximation achieved an LPPD of 1.0 and our Semi-Subspace approach achieves an LPPD of 1.14, while Semi-Subspace($k=12$) with an LPPD of 1.27 is even slightly better than \ac{hmc} (LPPD 1.26) running on the full parameter space. 

\subsection{Simulation Study}
\label{sec:exp_simu_study}
While simple, the previously analyzed data-generating process allows us to systematically investigate the behavior of our subspace approach in comparison to the gold standard full space \ac{hmc} method. Based on the previous results, we hypothesize that the subspace approach yields an unbiased estimate of the posterior mean, but is biased in the distribution's variance (and potentially higher moments). We examine this hypothesis by extending the data generation of our previous study, investigating two outcome distributions (Poisson and Normal), a larger input space $\vu \in \mathbb{R}^4$ and $\vx \in \mathbb{R}^3$, and different subspace dimensions $k={2,4,8,12,16}$. For every configuration, we conduct 50 simulation runs with different datasets. 
Each data point $(y_i, \vx_i, \vu_i)$ was generated using the following data generating process: First, we sample $\vu_i \sim N(0,\boldsymbol{I_4})$ and $\vx_i \sim N(0,\boldsymbol{I_3})$.
Next, we randomly initialize the SSR model with parameters $\vt^* \in \mathbb{R}^3$ and $\vw^* \in \mathbb{R}^{336}$, using a similar architecture as in our first experiment.
Finally we choose $y_i \sim \mathcal{N}(f(\vu_i) + \vx_i^\top\vt^*, 1)$ for the Normal outcome distribution case and $y_i \sim \Pois(\rho (f(\vu_i)+ \vx_i^\top\vt^*))$ with $\rho$ the exponential function for the  Poisson case. We model the synthetic data using an SSR model with the same architecture as in the data generation process.

\begin{figure}[htbp]
    \centering
    \includegraphics[width=0.95\linewidth]{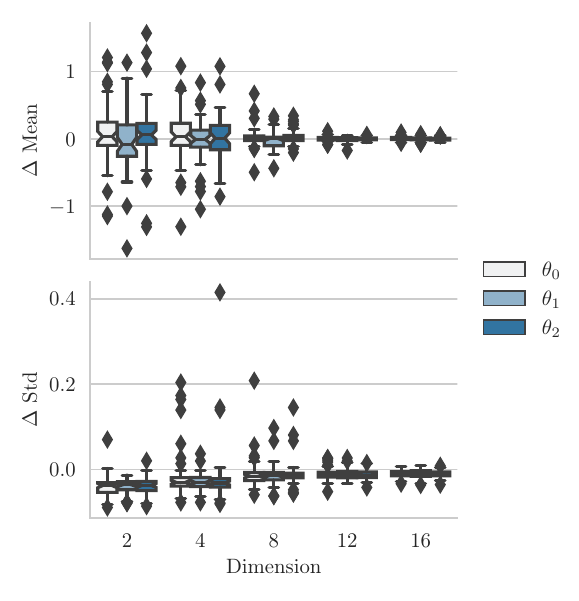} 
    \caption{Posterior mean (top) and standard deviation (bottom) of our approximation method compared to the gold standard \ac{hmc}. The boxplots show differences between the learned distribution's mean/standard deviation of our approach minus the respective statistic using \ac{hmc} for the 50 simulation repetitions. The x-axis depicts the different subspace dimensions $k$ used in our approach and each color represents one of the three parameters in $\theta$.}
    \label{fig:exp_simulation_statistics}
\end{figure}
\begin{figure}[htbp]
    \centering
    \includegraphics[width=0.75\columnwidth]{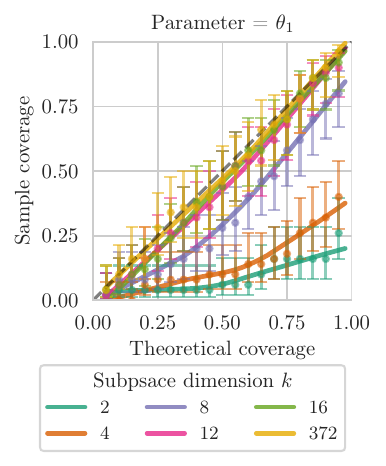} 
    \caption{Coverage comparison of credibility intervals derived from the posterior $p(\theta_1|\mathcal{D})$ using different subspace dimensions $k$ (colors). The theoretical coverage (x-axis) across different values in $(0,1)$ is plotted against the sample coverage (y-axis), based on the empirical ratio of the credibility interval containing the true parameter. 
    Whiskers represent the 95\% Wilson confidence interval.}
    \label{fig:exp_simulation_coverage}
\end{figure}

\textbf{Results}\quad 
We examine the results focusing on the differences in the mean and standard deviation of the posterior between our method and \ac{hmc}. Figure~\ref{fig:exp_simulation_statistics} shows the results for the Poisson distribution. We find that our subspace approach yields an unbiased posterior mean irrespective of $k$ and that the difference to \ac{hmc} reduces to zero for increasing subspace dimension (left plot). While biased in the distribution's variance, increasing the subspace alleviates this discrepancy and a larger subspace produces almost the same posterior variance as \ac{hmc}. Small $k$, in contrast, leads to overly confident uncertainty quantification.
In order to analyze previous results in light of parameter uncertainty quantification, we further check the calibration of the posterior. To this end, we compute the amount of coverage of the true parameter $\vt^*$ from the data generation process by the derived $\alpha$-credibility intervals for different nominal levels $\alpha\in(0,1)$. This is visualized by plotting the theoretical coverage $\alpha$ against the empirical sample coverage using the obtained posterior (cf.~Figure \ref{fig:exp_simulation_coverage} for $\theta_1$). Results clearly show that for increasing subspace dimensions, calibration improves, and already for $k=12$ or $k=16$, coverage is not notably different from the one obtained by using \ac{hmc} with 372 dimensions.  
\subsection{UCI Benchmark}
While the previous experiments assess our \ac{ssr} subspace approximation by checking the coverage of structured model parameters, we now evaluate the general applicability of our subspace approach using the benchmark datasets and methods investigated in \citet{wiese2022}. The authors show how to efficiently use multiple chains to capture different posterior modes and thereby achieve superior performance by capturing most of the relevant parts of the posterior. The obtained results are hence (close to) an oracle performance and can be used to check how well our approximation is working. Their benchmark comprises three simulated datasets and six datasets from the UCI machine learning repository \cite{dua2017uci}.
We use the same data splits, model architectures, and data pre-processing as in \citet{wiese2022}, allowing for a direct comparison with their \ac{mcmc} approach as well as results provided for Laplace approximation \citep{daxberger2021laplace} and Deep Ensembles \citep{lakshminarayanan2017simple}. We run our method using $k=2$ and $k=5$.

\textbf{Results}\quad Table~\ref{tab:UCI_large_net} summarizes the results, showing that our method outperforms both the Laplace approximation and deep ensembles on all provided datasets while being often very close to the gold standard \ac{mcmc} approach. We further see that an increase in subspace dimension can notably improve predictive performance (Diabetes, ForestF, Yacht). 
\begin{table*}[!h]
    \centering
    \small
        \caption{\small Normalized expected test log pointwise predictive density (LPPD; larger is better) with the network architecture introduced by \citet{wiese2022}, comprising three hidden layers with 16 neurons. The values within parentheses represent the standard errors of the predictive density per data point. The best method, excluding MCMC (representing an approximate upper bound), and all methods within one standard error of the best method are highlighted in bold.}
    \vskip 0.1in
    \begin{tabular}{l|r|rrrr}
    \toprule
    dataset & \ac{mcmc} & Subspace (k=2) & Subspace (k=5) & Deep Ens. & Laplace Appr. \\
    \midrule
    DI & 0.91 (±0.09)  & \textbf{0.82} (± 0.10) & \textbf{0.77} (± 0.09) & -2.02 (±0.02) & -1.81 (±0.01) \\
    DR & 0.95 (±0.08) & \textbf{0.82} (± 0.13) & \textbf{0.86} (± 0.12) & -2.20 (±0.02) & -2.33 (±0.00) \\
    Airfoil & 0.92 (±0.05) & \textbf{-0.28} (± 0.12) & \textbf{-0.19} (± 0.09) & -2.17 (±0.01) & -3.57 (±0.18) \\
    Concrete & 0.26 (±0.07) & \textbf{-0.53} (± 0.20) & \textbf{-0.55} (± 0.17) & -2.03 (±0.01) & -4.36 (±0.47) \\
    Diabetes & -1.18 (±0.08) & -2.40 (± 0.28) & \textbf{-1.21} (± 0.08) & -2.09 (±0.04) & -2.61 (±0.00) \\
    Energy & 2.07 (±0.46) & \textbf{1.43} (± 0.14) & \textbf{1.57} (± 0.15) & -1.99 (±0.02) & -1.39 (±0.06) \\
    ForestF & -1.43 (±0.45) & -1.90 (± 0.19) & \textbf{-1.38} (± 0.07) & -2.20 (±0.02) & -2.80 (±0.00) \\
    Yacht & 3.31 (±0.21) & -0.69 (± 1.90) & \textbf{1.49} (± 0.51) & -2.18 (±0.03) & -2.69 (±0.00) \\
    \bottomrule
    \end{tabular}
    \label{tab:UCI_large_net}
\end{table*}

\subsection{Application to Melanoma Data}
\label{sec:exp_sec_mela}

Finally, we apply our method to a real-world melanoma dataset \citep{isic2020} containing 33,058 patient records of $3\times 128\times 128$ RGB color images of skin lesions as well as additional metadata such as the patient's age. 
The primary objective of this dataset is to predict the presence or absence of malignant skin lesions. We follow the approach by \citet{durr_bernstein_2022} and process the images using a basic convolutional neural network (see Supplementary Material \ref{sec:app:exp_setup} for details) while modeling the patient's age as a linear effect $\theta_\text{age}$. Next to a comparison with \ac{mcmc} using only the age information, we compare against the transformation model approach by \citet{durr_bernstein_2022}, and the Laplace approximation (using the last layer approach). The data is split into six data folds as in previous works. 

\begin{figure}[htbp]
\centering
\includegraphics[width=\columnwidth]{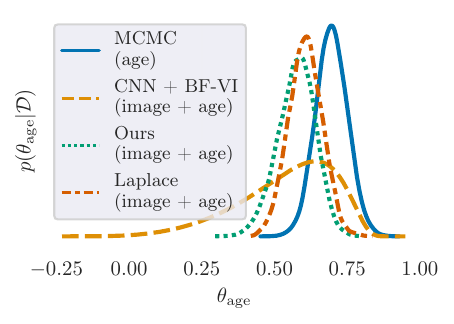}
\caption{Posterior $p(\theta_{\text{age}}|\mathcal{D})$ obtained using different methods on the Melanoma dataset. Distributions are based on a KDE smoother combining all samples from all training folds.}
\label{fig:melanoma_posterior}
\end{figure}

\textbf{Results}\quad In Figure~\ref{fig:melanoma_posterior}, we compare the posterior distribution $p(\theta_{\text{age}}|\mathcal{D})$ obtained by the different methods. 
Results suggest that the inclusion of image information decreases the effect of age when comparing the different methods to the results of \ac{mcmc} which only uses age information. 
The results from \citet{durr_bernstein_2022} (CNN + BF-VI) are not in line with all other methods, yielding a differently shaped distribution. While our approach is similar to the Laplace approximation in terms of the posterior $p(\theta_{\text{age}}|\mathcal{D})$ it surpasses both the Laplace approximation and the transformation model approach in terms of negative log-likelihood (see~Table~\ref{tab:Melanoma_performance}).
\begin{table}[]
    \centering
    \small
        \caption{\small Mean area under the ROC Curve (AUC) and LPPD values (standard errors in brackets; not available for CNN + BF-VI) across the six data folds for the different methods (rows)}
        \vskip 0.1in
         \resizebox{\columnwidth}{!}{
    \begin{tabular}{lcc}
    \toprule
     & AUC & LPPD \\
    \midrule
    CNN + BF-VI & 0.82 ($\pm$0.03) & -0.076 (NA) \\
    Laplace & 0.795 ($\pm$0.004) & -0.076 ($\pm$0.001) \\
    Semi-Subspace(k=2) & \textbf{0.841} ($\pm$0.003) & \textbf{-0.072} ($\pm$0.001) \\
    \bottomrule
    \end{tabular}    
    }
    \label{tab:Melanoma_performance}
\end{table}
\paragraph{Optimization Aspects}
It's worth noting that during our empirical analysis, we encountered difficulties fitting the Laplace \ac{ssr} model, requiring careful tuning of the learning rate, see Supplementary Material \ref{sec:app:add_results_mela} for a detailed discussion. 
We attribute this observation to the optimization asymmetry when training \ac{ssr} models, where the optimization of structured model parameters is not treated differently from the one of neural network parameters. 
%
We note that compared to the Laplace approximation, our method is less affected by the optimization asymmetry in \ac{ssr}, as we do not directly rely on a specific learning rate or optimizer (except for the construction of the sampling space). 
We find this to be a major advantage of our method compared to Laplace approximation or other optimization-based SSR methods.

\section{CONCLUSION}
We have presented a method to address two critical challenges inherent in SSR models, uncertainty quantification and optimization.
Our approach notably improves over na\"ive approximation methods in terms of posterior distribution quality while outperforming other approximation methods in predictive posterior performance. 
We also find that with a sufficiently large subspace dimension \( k \), our method comes remarkably close to replicating the posterior distribution and posterior predictive distribution achieved by gold standard \ac{mcmc} techniques.
Additionally, our method enables a deeper analysis of parameter uncertainty within the structured model component, accounting for the uncertainty propagated through the DNN part. This nuanced understanding of uncertainty provides valuable insights, particularly in domains like medical diagnostics, where model interpretability is crucial.
Furthermore, our work sheds light on the optimization asymmetry in \ac{ssr} and makes inference more robust as it mitigates the challenges arising from this asymmetry.
%
One limitation of our current method is the notable memory usage due to the storage of $(k+1) \times d$ parameters in $\boldsymbol{\Pi}$ and $\bar{\mathbf{p}}$ during the sampling phase, or in $\boldsymbol{\Lambda}$ during the training phase. A possible alleviation of this issue could be achieved by sequentially computing the necessary values in Equations \eqref{eq:bez} and \eqref{eq:sampling_projection}, trading performance with memory demand. 
Another limitation is the high computational demand, inherent in sampling-based approaches, requiring one forward pass per posterior sample. This could be addressed by treating only parts of the network as Bayesian. 

In summary, our method not only boosts the predictive performance of \ac{ssr} models but also serves as a comprehensive framework for understanding and quantifying uncertainty. It achieves results that are comparable to those of \ac{hmc}, while also addressing fitting challenges commonly encountered in other \ac{ssr} methods that rely solely on optimization. This makes our approach a valuable asset across a diverse array of applications.

\subsubsection*{Acknowledgements}
We thank the reviewers for their valuable feedback.
This work was supported by the Federal Ministry of Education and Research of Germany (BMBF) in the project “DeepDoubt” (grant no. 01IS19083A) and by the Carl-Zeiss-Stiftung in the project "DeepCarbPlanner".

\bibliography{bibliography}

\begin{thebibliography}{}

\bibitem[Amoura et~al., 2011]{amoura2011state}
Amoura, K., Wira, P., and Djennoune, S. (2011).
\newblock A state-space neural network for modeling dynamical nonlinear
  systems.
\newblock In {\em IJCCI (NCTA)}, pages 369--376.

\bibitem[Baumann et~al., 2021]{baumann.2020}
Baumann, P. F.~M., Hothorn, T., and {R{\"u}gamer}, D. (2021).
\newblock Deep conditional transformation models.
\newblock In {\em Machine Learning and Knowledge Discovery in Databases
  (ECML-PKDD)}, pages 3--18. Springer International Publishing.

\bibitem[Cheng et~al., 2016]{Cheng.2016}
Cheng, H.-T., Koc, L., Harmsen, J., Shaked, T., Chandra, T., Aradhye, H.,
  Anderson, G., Corrado, G., Chai, W., Ispir, M., et~al. (2016).
\newblock Wide \& deep learning for recommender systems.
\newblock In {\em Proceedings of the 1st workshop on deep learning for
  recommender systems}, pages 7--10. ACM.

\bibitem[Ciampi and Lechevallier, 1995]{ciampi1995designing}
Ciampi, A. and Lechevallier, Y. (1995).
\newblock Designing neural networks from statistical models: A new approach to
  data exploration.
\newblock In {\em KDD}, pages 45--50.

\bibitem[Ciampi and Lechevallier, 1997]{ciampi1997statistical}
Ciampi, A. and Lechevallier, Y. (1997).
\newblock Statistical models as building blocks of neural networks.
\newblock {\em Communications in statistics-theory and methods},
  26(4):991--1009.

\bibitem[Daxberger et~al., 2021]{daxberger2021laplace}
Daxberger, E., Kristiadi, A., Immer, A., Eschenhagen, R., Bauer, M., and
  Hennig, P. (2021).
\newblock Laplace redux-effortless bayesian deep learning.
\newblock {\em Advances in Neural Information Processing Systems},
  34:20089--20103.

\bibitem[de~Waal and du~Toit, 2007]{de2007generalized}
de~Waal, D.~A. and du~Toit, J.~V. (2007).
\newblock Generalized additive models from a neural network perspective.
\newblock In {\em Seventh IEEE International Conference on Data Mining
  Workshops (ICDMW 2007)}, pages 265--270. IEEE.

\bibitem[De~Waal and Du~Toit, 2011]{De.2011}
De~Waal, D.~A. and Du~Toit, J.~V. (2011).
\newblock Automation of generalized additive neural networks for predictive
  data mining.
\newblock {\em Applied Artificial Intelligence}, 25(5):380--425.

\bibitem[Dorigatti et~al., 2023]{dorigatti_frequentist_2023}
Dorigatti, E., Schubert, B., Bischl, B., and R\"ugamer, D. (2023).
\newblock Frequentist {{Uncertainty Quantification}} in {{Semi-Structured
  Neural Networks}}.
\newblock In {\em Proceedings of {{The}} 26th {{International Conference}} on
  {{Artificial Intelligence}} and {{Statistics}}}, pages 1924--1941. {PMLR}.

\bibitem[Dua et~al., 2017]{dua2017uci}
Dua, D., Graff, C., et~al. (2017).
\newblock Uci machine learning repository.

\bibitem[D{\"u}rr et~al., 2022]{durr_bernstein_2022}
D{\"u}rr, O., H{\"o}rling, S., Dold, D., Kovylov, I., and Sick, B. (2022).
\newblock Bernstein {{Flows}} for {{Flexible Posteriors}} in {{Variational
  Bayes}}.
\newblock {\em arXiv preprint arXiv:2202.05650}.

\bibitem[Garipov et~al., 2018]{garipov_loss_2018}
Garipov, T., Izmailov, P., Podoprikhin, D., Vetrov, D.~P., and Wilson, A.~G.
  (2018).
\newblock Loss {{Surfaces}}, {{Mode Connectivity}}, and {{Fast Ensembling}} of
  {{DNNs}}.
\newblock In {\em Advances in {{Neural Information Processing Systems}}},
  volume~31. {Curran Associates, Inc.}

\bibitem[Hubin et~al., 2018]{Hubin.2018}
Hubin, A., Storvik, G., and Frommlet, F. (2018).
\newblock {Deep Bayesian regression models}.

\bibitem[{International Skin Imaging Collaboration}, 2020]{isic2020}
{International Skin Imaging Collaboration} (2020).
\newblock Siim-isic 2020 challenge dataset.
\newblock Accessed on September 28, 2023.

\bibitem[Izmailov et~al., 2020]{izmailov_subspace_2020-1}
Izmailov, P., Maddox, W.~J., Kirichenko, P., Garipov, T., Vetrov, D., and
  Wilson, A.~G. (2020).
\newblock Subspace {{Inference}} for {{Bayesian Deep Learning}}.
\newblock In {\em Proceedings of {{The}} 35th {{Uncertainty}} in {{Artificial
  Intelligence Conference}}}, pages 1169--1179. {PMLR}.

\bibitem[Jantre et~al., 2023]{jantre2023learning}
Jantre, S., Urban, N.~M., Qian, X., and Yoon, B.-J. (2023).
\newblock Learning active subspaces for effective and scalable uncertainty
  quantification in deep neural networks.
\newblock {\em arXiv preprint arXiv:2309.03061}.

\bibitem[Kook et~al., 2022a]{kook_deep_2022}
Kook, L., G{\"o}tschi, A., Baumann, P.~F., Hothorn, T., and Sick, B. (2022a).
\newblock Deep interpretable ensembles.

\bibitem[Kook et~al., 2022b]{KOOK2022108263}
Kook, L., Herzog, L., Hothorn, T., Dürr, O., and Sick, B. (2022b).
\newblock Deep and interpretable regression models for ordinal outcomes.
\newblock {\em Pattern Recognition}, 122:108263.

\bibitem[Kopper et~al., 2022]{kopper2021}
Kopper, P., Wiegrebe, S., Bischl, B., Bender, A., and {R{\"u}gamer}, D. (2022).
\newblock {DeepPAMM: Deep Piecewise Exponential Additive Mixed Models for
  Complex Hazard Structures in Survival Analysis}.
\newblock In {\em Advances in Knowledge Discovery and Data Mining (PAKDD)},
  pages 249--261. Springer International Publishing.

\bibitem[Lakshminarayanan et~al., 2017]{lakshminarayanan2017simple}
Lakshminarayanan, B., Pritzel, A., and Blundell, C. (2017).
\newblock Simple and scalable predictive uncertainty estimation using deep
  ensembles.
\newblock {\em Advances in neural information processing systems}, 30.

\bibitem[Murray et~al., 2010]{murray2010elliptical}
Murray, I., Adams, R., and MacKay, D. (2010).
\newblock Elliptical slice sampling.
\newblock In {\em Proceedings of the thirteenth international conference on
  artificial intelligence and statistics}, pages 541--548. JMLR Workshop and
  Conference Proceedings.

\bibitem[Nelder and Wedderburn, 1972]{nelder1972generalized}
Nelder, J.~A. and Wedderburn, R.~W. (1972).
\newblock Generalized linear models.
\newblock {\em Journal of the Royal Statistical Society Series A: Statistics in
  Society}, 135(3):370--384.

\bibitem[Potts, 1999]{potts1999generalized}
Potts, W.~J. (1999).
\newblock Generalized additive neural networks.
\newblock In {\em Proceedings of the fifth ACM SIGKDD international conference
  on Knowledge discovery and data mining}, pages 194--200.

\bibitem[Pölsterl et~al., 2020]{Poelsterl.2020}
Pölsterl, S., Sarasua, I., Gutiérrez-Becker, B., and Wachinger, C. (2020).
\newblock A wide and deep neural network for survival analysis from anatomical
  shape and tabular clinical data.
\newblock {\em Communications in Computer and Information Science}, page
  453–464.

\bibitem[R\"{u}gamer, 2023]{pho}
R\"{u}gamer, D. (2023).
\newblock A new {PHO}-rmula for improved performance of semi-structured
  networks.
\newblock In {\em Proceedings of the 40th International Conference on Machine
  Learning}, volume 202 of {\em Proceedings of Machine Learning Research},
  pages 29291--29305. PMLR.

\bibitem[R{\"u}gamer et~al., 2023]{rugamer_semi-structured_2022}
R{\"u}gamer, D., Kolb, C., and Klein, N. (2023).
\newblock Semi-structured distributional regression.
\newblock {\em The American Statistician}, 0(0):1--12.

\bibitem[Sick et~al., 2021]{sick2021deep}
Sick, B., Hothorn, T., and D{\"u}rr, O. (2021).
\newblock Deep transformation models: Tackling complex regression problems with
  neural network based transformation models.
\newblock In {\em 2020 25th International Conference on Pattern Recognition
  (ICPR)}, pages 2476--2481. IEEE.

\bibitem[{Tran} et~al., 2020]{Tran.2018}
{Tran}, M.-N., {Nguyen}, N., {Nott}, D., and {Kohn}, R. (2020).
\newblock Bayesian deep net {GLM} and {GLMM}.
\newblock {\em Journal of Computational and Graphical Statistics},
  29(1):97--113.

\bibitem[Wahba, 1990]{wahba1990spline}
Wahba, G. (1990).
\newblock {\em Spline models for observational data}.
\newblock SIAM.

\bibitem[Wiese et~al., 2023]{wiese2022}
Wiese, J.~G., Wimmer, L., Papamarkou, T., Bischl, B., G\"unnemann, S., and
  R{\"u}gamer, D. (2023).
\newblock Towards efficient posterior sampling in deep neural networks via
  symmetry removal.
\newblock In {\em Machine Learning and Knowledge Discovery in Databases
  (ECML-PKDD)}. Springer International Publishing.

\bibitem[Wood, 2017]{wood2017generalized}
Wood, S.~N. (2017).
\newblock {\em Generalized additive models: an introduction with R}.
\newblock CRC press.

\bibitem[Wortsman et~al., 2021]{wortsman2021learning}
Wortsman, M., Horton, M.~C., Guestrin, C., Farhadi, A., and Rastegari, M.
  (2021).
\newblock Learning neural network subspaces.
\newblock In {\em International Conference on Machine Learning}, pages
  11217--11227. PMLR.

\bibitem[Zhang et~al., 2017]{zhang2017understanding}
Zhang, C., Bengio, S., Hardt, M., Recht, B., and Vinyals, O. (2017).
\newblock Understanding deep learning requires rethinking generalization.
\newblock In {\em International Conference on Learning Representations}.

\bibitem[Zhang et~al., 2021]{zhang2021understanding}
Zhang, C., Bengio, S., Hardt, M., Recht, B., and Vinyals, O. (2021).
\newblock Understanding deep learning (still) requires rethinking
  generalization.
\newblock {\em Communications of the ACM}, 64(3):107--115.

\end{thebibliography}

\appendix
\setcounter{section}{0} 
\onecolumn
\aistatstitle{Bayesian Semi-structured Subspace Inference: \\
Supplementary Materials}
\section{TEMPERED POSTERIOR FOR SSR MODELS}
\label{app:temp}
In the following section, we will discuss how we can apply a temperature parameter to improve predictive performance and what the unique challenges are in the context of SSR models.

As highlighted by \citet{izmailov_subspace_2020-1}, subspace inference can yield overly confident uncertainty estimates. 
This overconfidence may stem from the fact that the prior is defined within the subspace of dimension $k$ rather than the larger dimension $d$. 
Consequently, reducing the parameter space through subspace construction has a noticeable impact on the posterior distribution. 
To mitigate this effect, \citet{izmailov_subspace_2020-1} proposed the application of a temperature parameter $T > 0$. 
This parameter scales the likelihood according to the following Equation:

\begin{equation}
p_T(\vt, \vphi|\mathcal{D}) \propto p(\mathcal{D}|\vt, \vphi)^{1/T} p(\vt, \vphi)
\label{eq:likelihood_semi-subspace_temp}
\end{equation}

Here, a temperature smaller than one shifts the posterior towards the maximum likelihood estimate, while a temperature larger than one moves the posterior closer to the prior distribution. 
Their findings suggest that using a temperature parameter can improve predictive performance and, potentially, the quality of uncertainty estimates.

However, with this proposed method the marginal posterior distribution $p(\vt|\mathcal{D})$ will be influenced by the temperature parameter.
This is problematic for SSR models as $\vt$ is not affected by the subspace approximation, hence there is no reason to modify its marginal posterior with a temperature parameter.
To tackle this challenge, we devised a novel approach.
First, we split the joint posterior distribution into two parts, which can be expressed as follows:
\begin{equation}
\begin{aligned}
    p_T(\vt, \vphi|\mathcal{D}) &= p(\vt| \vphi, \mathcal{D})\:p_T(\vphi|\mathcal{D}) \\
    &= \frac{p(\mathcal{D}|\vt, \vphi)\:p(\vt | \vphi)}{p(\mathcal{D}|\vphi)} \: \frac{p(\mathcal{D}|\vphi)^{\frac{1}{T}}\:p(\vphi)}{p_T(\mathcal{D})}
\label{eq:post_condition_temperature}
\end{aligned}
\end{equation}
The first part of this equation represents the conditioned posterior for our interpretable parameters, denoted as $\vt$, while the second part reflects the posterior of the neural network, $p(\vphi|\mathcal{D})$, where we apply the temperature parameter.
If $p(\vt|\mathcal{D})$ and $p(\vphi|\mathcal{D})$ are independent, the temperature parameter won't influence the posterior $p(\vt|\mathcal{D})$ in this approach. This can be shown by simply rewriting Equation~\eqref{eq:post_single_parameter_contin}
\begin{flalign*}
&& p_T(\vt|\mathcal{D}) &= \int p_T(\vt, \vphi|\mathcal{D}) d\vphi \\
&&                      &= \int p(\vt|\vphi, \mathcal{D})\:p_T(\vphi|\mathcal{D}) d\vphi \\
&&                      &= p(\vt| \mathcal{D})\:\int p_T(\vphi|\mathcal{D}) d\vphi  &, \text{if} \, p(\vt, \vphi|\mathcal{D}) = p(\vt|\mathcal{D})p(\vphi|\mathcal{D})\\
&&                      &= p(\vt| \mathcal{D}) \qquad \blacksquare \refstepcounter{equation}\tag{\theequation}
\label{eq:post_single_parameter_contin}
\end{flalign*}

However, the proposed approach necessitates the computation of the marginal likelihood, represented as $p(\mathcal{D}|\vphi)$. 
We chose to numerically integrate the likelihood function according to the following equation:
\begin{equation}
p(\mathcal{D}|\vphi) = \int{p(\mathcal{D}|\vt,\vphi) \: p(\vt|\vphi) \: d\vt}
\label{eq:integral_likelihood}
\end{equation}
It's important to note that this numerical integration depends on the choice of technical parameters like integration step size and range, making misconfigurations of these parameters a potential source of inaccurate results.
Additionally, this approach is only feasible when the dimension of the structural parameters is relatively small, which allows for manageable numerical integration. 
Given these constraints, applying a temperature parameter to SSR models proves to be a challenging endeavor.

In our initial analysis, we applied our proposed temperature adjustment to the likelihood function of an SSR model according to Equation~\ref{eq:post_condition_temperature}.
We conducted this analysis on the melanoma dataset, where the parameter $\theta_{\text{age}}$ is of low dimension (1-dimensional) which is crucial to perform numerical integration. 
In this experiment, we observed that the posterior $p(\theta_{\text{age}}|\mathcal{D})$ of interest is only slightly influenced by the temperature parameter.
While we were able to achieve marginal performance improvements, these minimal gains are outweighed by the associated computational challenges and the temperature's influence on the posterior $p(\theta_{\text{age}}|\mathcal{D})$.
Consequently, implementing a tempered posterior for SSR models in practice is challenging, leading us to decide against its adoption.

\section{ADDITIONAL PROOFS}
\subsection{The B\'ezier Curve Resides in the Affine Subspace}
\label{proof:bezier}
We aim to demonstrate that every point on the B\'ezier curve \(B_{\lambda}(t)\), which serves as an approximation of the low-energy valley, can be obtained through our chosen sampling procedure. Specifically, our sampling approach ensures that each sample \(\vphi\) belongs to the affine subspace spanned by the control points \(\vp_0^*, \vp_1^*, \ldots, \vp_k^*\).
\begin{equation}
\text{AffSpan}(\boldsymbol{\Lambda}^*) = \left\{ \vp_0^* + \sum_{i=1}^{k} \varphi'_i \Delta \vp^*_i \,\Big| \, \varphi'_i \in \mathbb{R} \right\}.
\end{equation}
To demonstrate that the B\'ezier curve \( B_{\lambda}(t) \) resides within this affine subspace, we perform the following algebraic manipulations:
\begin{align*}
    B_{\lambda}(t) &= \sum_{l=0}^{k} \binom{k}{l} (1-t)^{k-l} t^l \vp_l^* \\
    &= (1-t)^k \vp_0^* + \sum_{l=1}^{k} \binom{k}{l} (1-t)^{k-l} t^l \vp_l^* \\
    &= (1-t)^k \vp_0^* + \sum_{l=1}^{k} \binom{k}{l} (1-t)^{k-l} t^l (\vp_l^* - \vp_0^* + \vp_0^*) \\
    &= \vp_0^* \cdot \underbrace{\left( \sum_{l=0}^{k} \binom{k}{l} (1-t)^{k-l} t^l \right)}_{=1} + \sum_{l=1}^{k} \underbrace{\binom{k}{l} (1-t)^{k-l} t^l}_{=\alpha_l} (\vp_l^* - \vp_0^*) \\
    &= \vp_0^* + \sum_{l=1}^{k} \alpha_l (\vp_l^* - \vp_0^*), 
\end{align*}
where \( \alpha_l = \binom{k}{l} (1-t)^{k-l} t^l \) for \( l \) in \( 1, \ldots, k \).
The last equation clearly indicates that \( B_{\lambda}(t) \) is contained within the affine subspace \(\text{AffSpan}(\boldsymbol{\Lambda}^*)\).

\section{EXPERIMENTAL SETUP}
\label{sec:app:exp_setup}

The code and notebooks with further experimental settings for this project are available in an anonymous repository under \href{https://anonymous.4open.science/r/Bayesian_Semi_Sub-8655}{https://anonymous.4open.science/r/Bayesian\_Semi\_Sub-8655}. It will be made public upon submission. 

\subsection{General Experimental Setup}
Below, we detail how we train the model to construct the subspace. 
To train the model, we optimized the entire SSR model, where only the weights of the DNN part are controlled via the Bézier curve. 
In all experiments, we used the Adam optimizer with a learning rate of 1e-4 and weight decay of 1e-4 for the medical dataset, a learning rate of 5e-3 with zero weight decay for the UCI benchmark, and a learning rate of 0.0025 and weight decay of 1e-3 for the Simulation and toy data.
After a fixed number of epochs, we selected the model with the lowest validation loss and used this model to construct the subspace.
We also verified that the training was long enough for the model to enter the overfitting region and adapted the number of epochs accordingly.

\subsection{CNN Model Architecture}

The following listing defines the PyTorch model which we used to process the medical dataset in  Experiment \ref{sec:exp_sec_mela}. 
The following SSR model architecture consists of a CNN (DNN listing) and a linear structured model part (structured\_model listing). 
\begin{lstlisting}[caption={DNN definition used for the melanoma data in Section~\ref{sec:exp_sec_mela} }]
(DNN): Sequential(
    (0): Sequential(
        (0): Conv2d(3, 32, kernel_size=(3, 3), stride=(1, 1))
        (1): Tanh()
        (2): MaxPool2d(kernel_size=2, stride=2, padding=0, 
                       dilation=1, ceil_mode=False)
        (3): Conv2d(32, 32, kernel_size=(3, 3), stride=(1, 1))
        (4): Tanh()
        (5): MaxPool2d(kernel_size=2, stride=2, padding=0, 
                       dilation=1, ceil_mode=False)
        (6): Conv2d(32, 64, kernel_size=(3, 3), stride=(1, 1))
        (7): Tanh()
        (8): MaxPool2d(kernel_size=2, stride=2, padding=0, 
                       dilation=1, ceil_mode=False)
        (9): Conv2d(64, 64, kernel_size=(3, 3), stride=(1, 1))
        (10): Tanh()
        (11): MaxPool2d(kernel_size=2, stride=2, padding=0, 
                       dilation=1, ceil_mode=False)
        (12): Conv2d(64, 128, kernel_size=(3, 3), stride=(1, 1))
        (13): Tanh()
        (14): MaxPool2d(kernel_size=2, stride=2, padding=0, 
                       dilation=1, ceil_mode=False)
    )
    (1): Sequential(
        (0): Flatten(start_dim=1, end_dim=-1)
        (1): Linear(in_features=512, out_features=128, bias=True)
        (2): Tanh()
        (3): Linear(in_features=128, out_features=128, bias=True)
        (4): Tanh()
        (5): Linear(in_features=128, out_features=1, bias=True)
    )
)
(structured_model): Linear(in_features=1, out_features=1, bias=False)
\end{lstlisting}

\section{ADDITIONAL RESULTS}
\label{sec:app:add_results}

In this section, we provide further results from our experiments.

\subsection{Additional Results from the Toy Data}
\label{sec:app:add_results_toydata}

\begin{figure}[htbp]
    \centering
    \begin{minipage}[c]{0.48\linewidth}
    \centering
    \includegraphics[width=\linewidth]{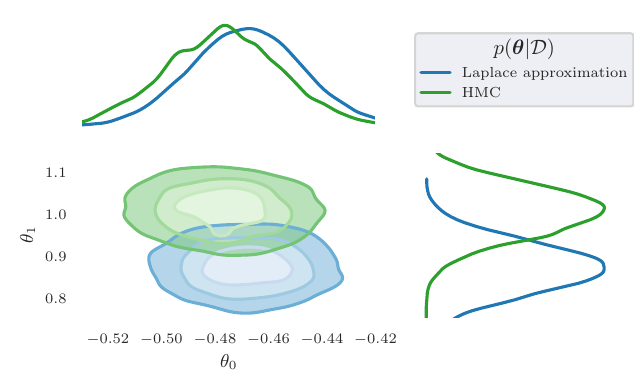} 
    \caption{Comparison of the posterior distribution $p(\vt|\mathcal{D})$ between ground truth HMC and Laplace Approximation on the regression dataset adapted from \cite{izmailov_subspace_2020-1}. This posterior distribution corresponds to the posterior predictive shown in Figure~\ref{fig:exp_toy_post_pred_Laplace} }
    \label{fig:exp_toy_post_comp_Laplace}
    \end{minipage}
    \hfill
    \begin{minipage}[c]{0.48\linewidth}
    \centering
    \includegraphics[width=\linewidth]{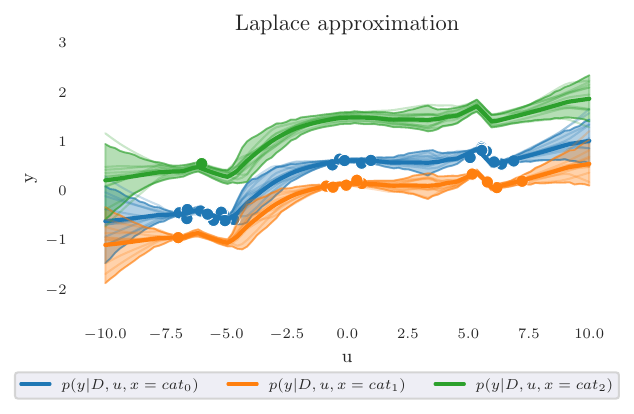} 
    \caption{Posterior predictive using the Laplace approximation on the toy dataset. We applied the Laplace approximation on the last layer of the DNN part and on the two-dimensional $\vt$ parameters of the structured model part. Data and model architecture are the same as described in Section~\ref{sec:exp_naive_sub}}. 
    \label{fig:exp_toy_post_pred_Laplace}
    \end{minipage}
\end{figure}

Here, we present additional results for the na\"ive Laplace approximation on the toy dataset.
Figure~\ref{fig:exp_toy_post_comp_Laplace} shows the posterior $p(\vt|\mathcal{D})$ and Figure~\ref{fig:exp_toy_post_pred_Laplace} the corresponding posterior predictive distribution obtained by applying the na\"ive Laplace approximation to the last layer and the structured model component. 
In the posterior distribution, we see some deviations between na\"ive Laplace approximation and HMC. 
However, the posterior predictive distribution is comparable to our approach with a subspace dimension of two.
If we compare to HMC or our method with a subspace dimension of 12 it is still too narrow in terms of epistemic uncertainty (See Figure~\ref{fig:fig1}). 

\begin{figure}[htbp]
    \begin{minipage}[c]{0.48\linewidth}
    \centering
    \includegraphics[width=\linewidth]{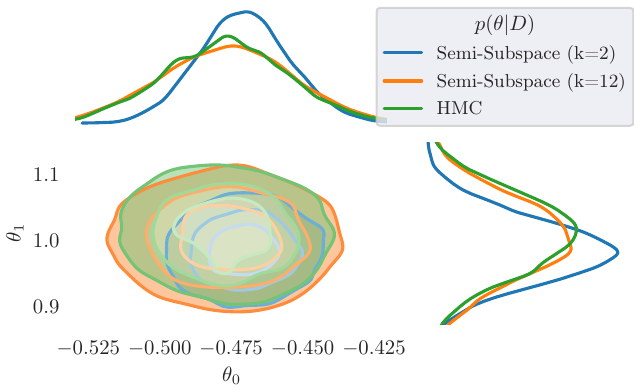} 
    \caption{Posterior of the parameters in the structured model part using our approach with $k=2$ and $k=12$ (Semi-Subspace) compared with \ac{hmc} running in the full parameter space. The top and right plots show the marginal posterior distribution, whereas the center plot visualizes the bivariate distribution using a kernel density plot based on 4000 samples from 10 \ac{hmc} chains.}
    \label{fig:exp_toy_additional_post_comp}
    \end{minipage}
    \hfill
\end{figure}

To complement the experiment on the toy dataset, we present additional results in the posterior distribution of the parameters from the structured model component in Figure~\ref{fig:exp_toy_additional_post_comp}. 
The results for HMC and Semi-Subspace ($k=2$) are identical to those shown in Figure~\ref{fig:naive_post_compare}, and we extended the results with our Semi-Subspace model, utilizing a subspace dimension of 12. 
The corresponding posterior predictive was shown in Figure~\ref{fig:fig1}. 
This comparison aligns with the findings from our simulation study, indicating that increasing the subspace dimension leads to posterior distributions that closely resemble those obtained with HMC.

\FloatBarrier
\newpage

\subsection{Additional Results from the Simulation Study}
\begin{figure}[ht]
    \centering
    \includegraphics[width=\linewidth]{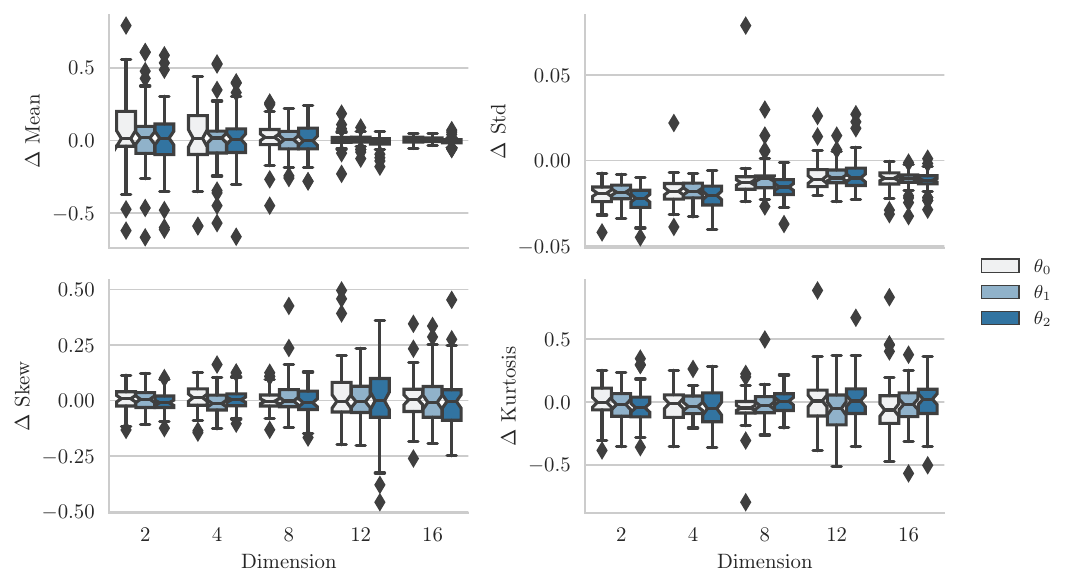} 
    \caption{A detailed comparison of the differences in the first four moments between our approximation method and the gold standard HMC when utilizing a normal outcome distribution in our simulation study. The boxplots show differences between the learned distribution’s moment of our approach minus the respective moment using HMC for the 50 simulation repetitions. The x-axis depicts the different subspace dimensions $k$ used in our approach and each color represents one of the three parameters in $\vt$.}
    \label{fig:exp_simulation_stats_normal_all}
\end{figure}

We continue with further results of our simulation study. Figure~\ref{fig:exp_simulation_stats_normal_all} and Figure~\ref{fig:exp_simulation_coverage_normal_all_params} show the results by using a normal outcome distribution, where the parameters $\mu$ is modeled, instead of the Poisson distribution. 
These Figures also validate our thesis, that increasing the subspace dimension reduces the error in the first two moments of the posterior $p(\vt|\mathcal{D})$ and improves its uncertainty quality.

In Figure~\ref{fig:exp_simulation_coverage} we visualized the influence of the subspace dimensions on the posterior calibration. This was shown by depicting one parameter $\theta_1$ out of the three-dimensional parameter space. 
For the sake of completeness, we provide in the following Figure~\ref{fig:exp_simulation_coverage_poisson_all_params} the calibration comparison for the entire parameter space $\vt$ from the structured model.
In a parallel analysis to the previous one, Figure~\ref{fig:exp_simulation_coverage_normal_all_params} showcases the calibration comparison using the Normal outcome distribution.
\begin{figure}[!ht]
    \centering
    \includegraphics[width=\linewidth]{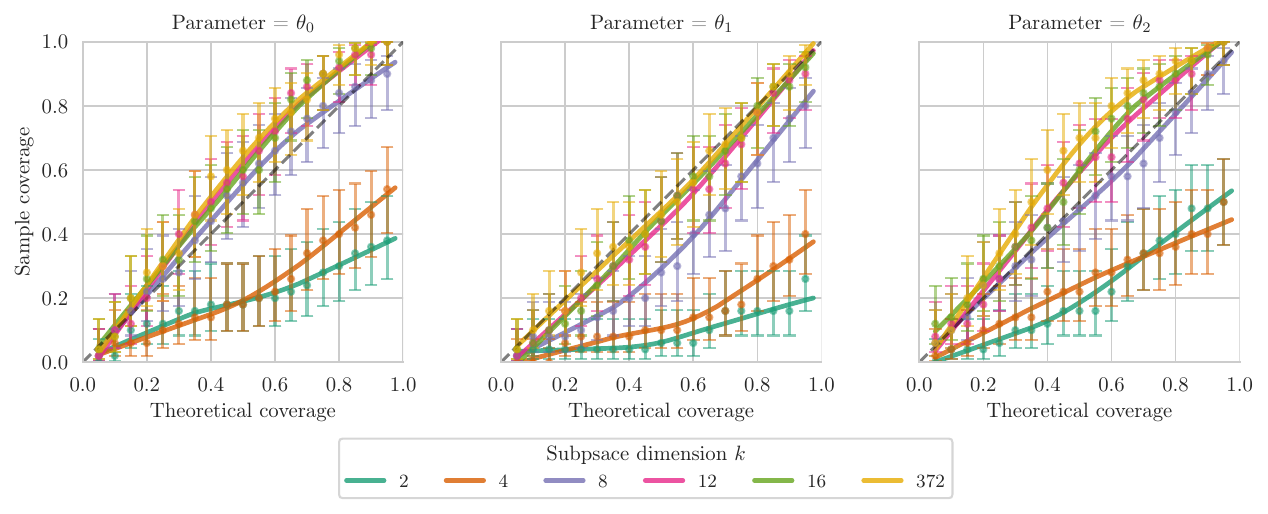}
    \caption{Coverage comparison of credibility intervals derived from the posterior $p(\theta_1|\mathcal{D})$ using different subspace dimensions $k$ (colors) of all parameters instead of only picking $\theta_1$ as shown in Figure~\ref{fig:exp_simulation_coverage}. The theoretical coverage (x-axis) across different values in $(0,1)$ is plotted against the sample coverage (y-axis), based on the empirical ratio of the credibility interval containing the true parameter. 
    Whiskers represent the 95\% Wilson confidence interval.}
    \label{fig:exp_simulation_coverage_poisson_all_params}
\end{figure}

\begin{figure}[!ht]
    \centering
    \includegraphics[width=\linewidth]{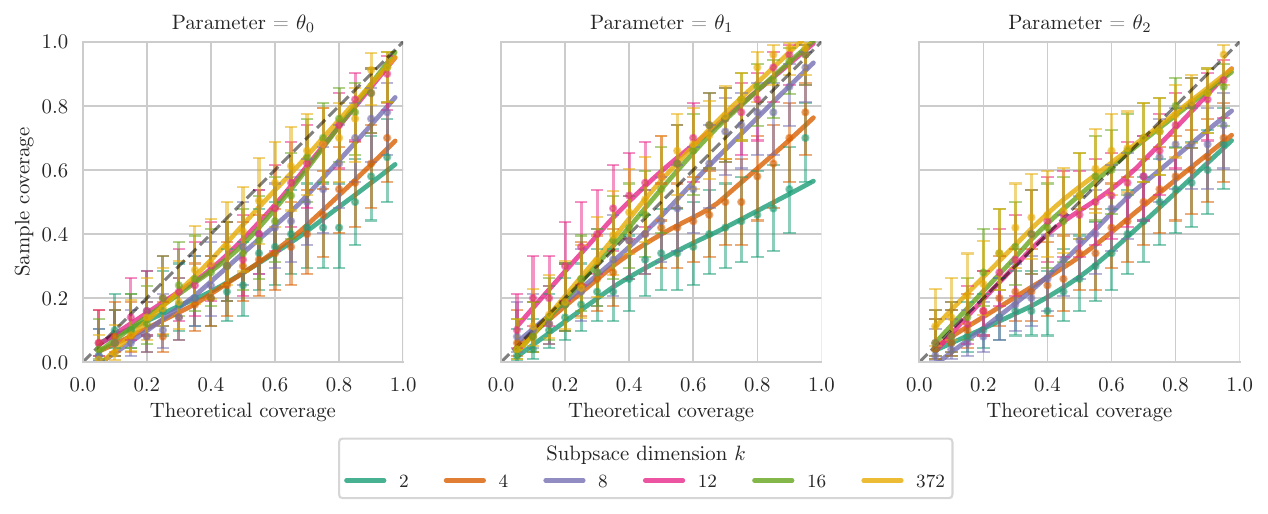} 
    \caption{Identical analyses as in Figure~\ref{fig:exp_simulation_coverage_poisson_all_params} but with Normal outcome distribution instead of Poisson distribution}
    \label{fig:exp_simulation_coverage_normal_all_params}
\end{figure}

\FloatBarrier

\newpage
\subsection{Additional Results from the Melanoma Dataset}
\label{sec:app:add_results_mela}
\begin{figure}[ht]
    \centering
    \includegraphics[width=\linewidth]{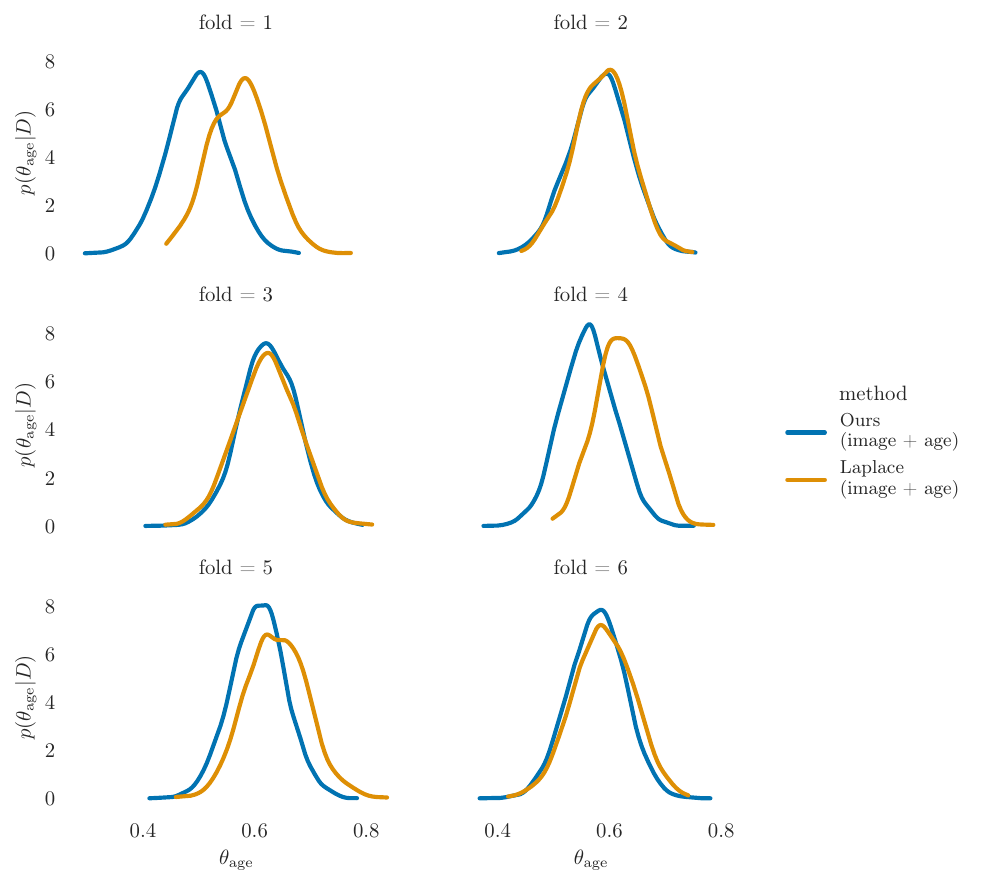} 
    \caption{Posterior $p(\theta_{\text{age}}|\mathcal{D})$ for each of the six folds in the melanoma dataset. The color-coding distinguishes between our method and the Laplace approximation (using a learning rate of 5e-3). The corresponding loss curve is displayed in Figure~\ref{fig:exp_melanoma_Laplace_loss_curve}. According to this loss curve, we select the model with the lowest validation loss and accomplished the Laplace approximation to generate the shown posterior.
    }
    \label{fig:exp_melanoma_post_per_split}
\end{figure}

In Figure~\ref{fig:melanoma_posterior}, we present the overall posterior $p(\theta_{\text{age}}|\mathcal{D})$ by pooling all samples from the six folds. 
For a more detailed examination, we break down the analysis in Figure~\ref{fig:exp_melanoma_post_per_split} to visualize the posterior distribution separately for each fold, rather than aggregating the samples. 
We observe that in some folds (2, 3, and 6), the posterior distribution of the Laplace approximation closely resembles our approach. 
However, folds one and four exhibit slight differences, but we did not observe a consistent trend. 
Notably, the expectation of the posterior remains relatively stable, with fold one being the exception. This suggests that pooling the folds, as shown in Figure~\ref{fig:melanoma_posterior}, does not significantly alter the distribution's shape.

\newpage
\paragraph{Optimization Asymmetry}

In the following discussion, we investigate the optimization asymmetry present in SSR models which are optimized using gradient-based methods.  
Optimization asymmetry, as we define it, suggests that the parameters $\vt$ of the structured model part require different optimization strategies compared to the weights $\vw$ of a DNN.

\begin{figure}[ht]
    \centering
    \begin{minipage}[c]{0.48\linewidth}
        \centering
        \includegraphics[width=\linewidth]{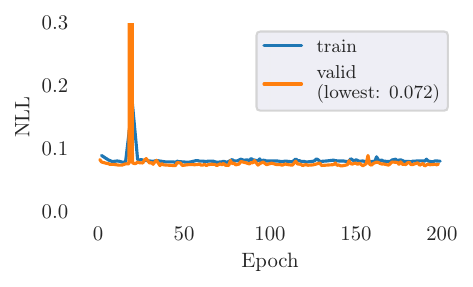} 
        \caption{Validation and training negative log-likelihood (NLL) (Lower is better). This loss curve corresponds to the Laplace approximation training on fold 3 with the larger learning rate 5e-3}
        \label{fig:exp_melanoma_Laplace_loss_curve}
    \end{minipage}
    \hfill
    \begin{minipage}[c]{0.48\linewidth}
        \centering
        \includegraphics[width=\linewidth]{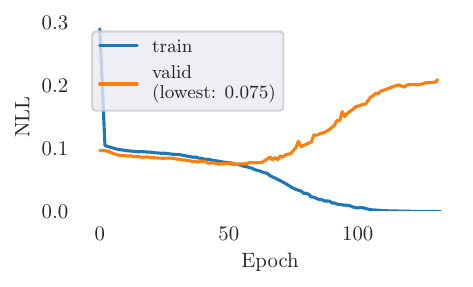} 
        \caption{Same validation and training negative log-likelihood as shown in Figure~\ref{fig:exp_melanoma_Laplace_loss_curve} while learning rate for the \ac{ssr} model was 1e-4 compared to 5e-3}
        \label{fig:exp_melanoma_Laplace_loss_curve_small_lr}
    \end{minipage}
\end{figure}

To investigate this issue we trained two SSR models, where each one was trained with a different learning rate on the melanoma dataset.
In the first experiment, we trained a model with a learning rate of 5e-3, as utilized in our primary work. 
Analyzing the corresponding validation loss curve (refer to Figure~\ref{fig:exp_melanoma_Laplace_loss_curve}), one could argue for a decrease in the learning rate due to a peak and fast optimization in the early optimization phases.
Thus, we tested a lower learning rate of 1e-4. 
This leads to a smoother loss curve and is, therefore, more trustworthy at first sight (as shown in Figure~\ref{fig:exp_melanoma_Laplace_loss_curve_small_lr}). 
However, the lower learning rate shows an optimization issue for the parameter $\theta_{\text{age}}$ in the structured model part.

To highlight this issue, we take a look at the posterior $p(\theta_{\text{age}}|\mathcal{D})$ and compare
the posterior approximation of our approach against the Laplace approximation obtained from the trained SSR model which could be affected by optimization issues.
We repeated the training over the six folds from the melanoma dataset with different weight initialization. As in all experiments, we selected the model with the lowest validation loss to perform the Laplace approximation.

\begin{figure}[ht]
    \centering
    \includegraphics[width=\linewidth]{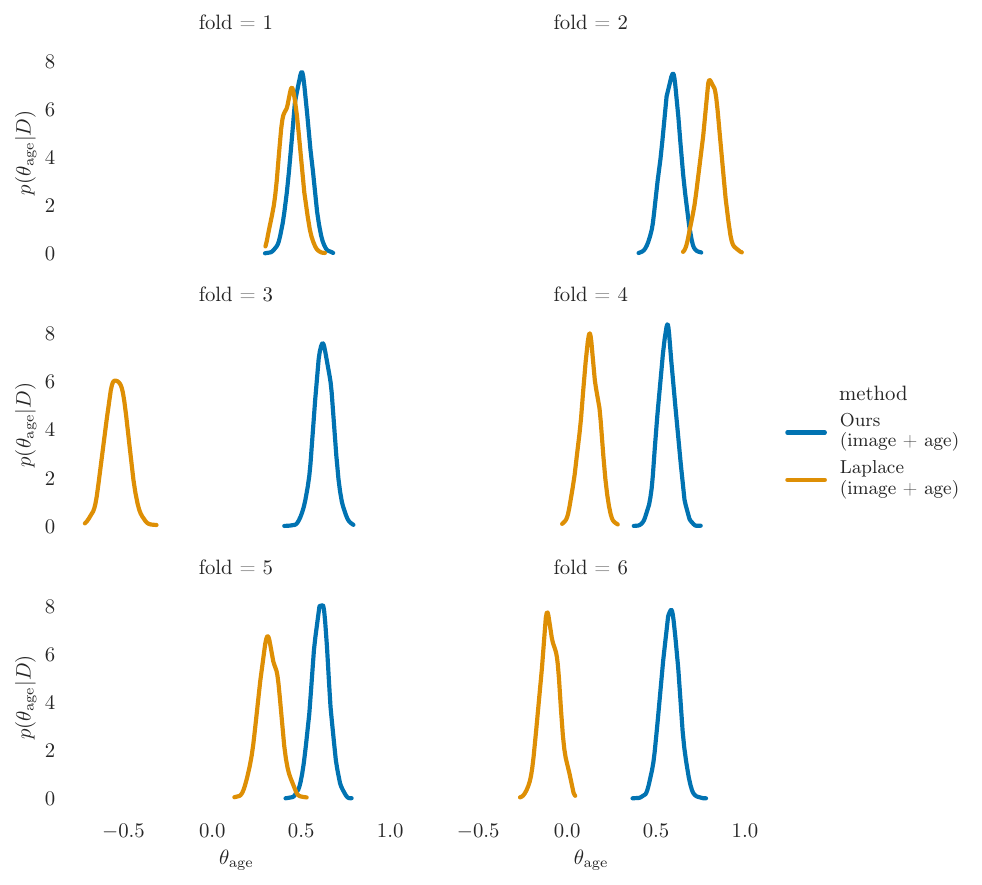} 
    \caption{Same analysis as shown in Figure~\ref{fig:exp_melanoma_post_per_split}, but the Laplace approximation model is trained with a smaller learning rate (1e-4). However, according to the validation loss (cf. Figure~\ref{fig:exp_melanoma_Laplace_loss_curve_small_lr}), the Laplace approximation model was still able to achieve overfitting}
    \label{fig:exp_melanoma_post_per_split_low_lr}
\end{figure}

Figure~\ref{fig:exp_melanoma_post_per_split} displays the results from our primary work, which employed the larger learning rate, while Figure~\ref{fig:exp_melanoma_post_per_split_low_lr} illustrates the posterior $p(\theta_{\text{age}}|\mathcal{D})$ using the lower learning rate.
In Figure~\ref{fig:exp_melanoma_post_per_split_low_lr} we see the optimization issue because the Laplace approximation exhibits significantly higher variance in the expectation of the posterior across the six folds compared to our approach and the results shown in Figure~\ref{fig:exp_melanoma_post_per_split}.

We argue that this asymmetry arises because the change a parameter must undergo from initialization to maximum likelihood solution is typically much larger for $\vt$ compared to $\vw$. 
Our intuition behind this is, that it exists an arbitrary number of solutions for $\vw$, including those closer to the initialization point. 
Therefore, optimizing $\vt$ may necessitate a greater number of optimization steps or larger learning rates compared to $\vw$.
We observed this in our experiments as the expectation of $p(\theta_{\text{age}}|\mathcal{D})$ still show changes   
when optimization is finished according to the validation loss.
Thus, in the epoch where we stop training due to the signs of overfitting, the parameter $\theta_{\text{age}}$ of the structured model part still shows a dependence on the initial value.
So while a lower learning rate produces a smoother learning curve, the increased learning rate leads to more stable results for the structured part (cf. Figure~\ref{fig:exp_melanoma_post_per_split} vs. \ref{fig:exp_melanoma_post_per_split_low_lr}).  


This highlights the challenges of optimizing SSR models, which we attribute to what we refer to as \textit{optimization asymmetry}.



\FloatBarrier
\newpage
\subsection{Additional Results with the smaller Network Architecture from \cite{wiese2022} on the UCI Benchmark}
For the sake of completeness, we also conducted our benchmark study using the 'smaller network f1' proposed by \cite{wiese2022} on the simulated and UCI datasets. 
Table~\ref{tab:UCI_small_net} confirms our previous results, showing that increasing the subspace dimension improves predictive performance. 
Additionally, with this network architecture, our approach is capable of achieving performance close to HMC on almost every dataset, and it outperforms in most cases the other two approximation methods.
\begin{table*}[!h]
    \centering
    \small
        \caption{\small Normalized expected test log pointwise predictive density (LPPD; larger is better) with the "smaller network f1" introduced by \citet{wiese2022}, comprising a single hidden layer with three neurons. The values within parentheses represent the standard errors of the predictive density per data point. The best method, excluding MCMC (representing an approximate upper bound), and all methods within one standard error of the best method are highlighted in bold.
        }
        \vskip 0.1in
    \resizebox{0.95\textwidth}{!}{
    \begin{tabular}{l|r|rrrrr}
    \toprule
    dataset & MCMC & Subspace (k=2) & Subspace (k=5) & Deep Ens. & Laplace Appr. \\
    \midrule
    DS & -0.53 (±0.09) & \textbf{-0.58 (± 0.11)} & \textbf{-0.60 (± 0.11)} & \textbf{-0.58 (±0.11)} & \textbf{-0.57 (±0.10)} \\
    DI & 0.79 (±0.06) & 0.51 (± 0.06) & \textbf{0.60 (± 0.05)} & \textbf{0.56 (±0.06)} & 0.53 (±0.07) \\
    DR & 0.64 (±0.10) & -0.39 (± 0.11) &\textbf{0.57 (± 0.12)} & -1.46 (±0.06) & -27.39 (±3.65) \\
    Airfoil & -0.74 (±0.04) & \textbf{-0.88 (± 0.05)} & \textbf{-0.83 (± 0.06)} & -1.62 (±0.03) & -1.78 (±0.13) \\
    Concrete & -0.41 (±0.05) & \textbf{-0.53 (± 0.06)} & \textbf{-0.50 (± 0.06)} & -1.59 (±0.03) & -14.49 (±1.02) \\
    Diabetes & -1.20 (±0.07) & -1.24 (± 0.09) & \textbf{-1.17 (± 0.06)} & -1.47 (±0.07) & -1.46 (±0.09) \\
    Energy & 0.92 (±0.04) & -0.05 (± 0.07) & \textbf{0.62 (± 0.08)} & -1.76 (±0.02) & -31.74 (±1.88) \\
    ForestF & -1.37 (±0.07) & -1.47 (± 0.08) & \textbf{-1.37 (± 0.07)} & -1.60 (±0.06) & -2.39 (±0.16) \\
    Yacht & 1.90 (±0.16) & \textbf{1.13 (± 0.47)} & \textbf{1.20 (± 0.44)} & -1.14 (±0.14) & -5.60 (±1.51) \\
    \bottomrule
    \end{tabular}
    }
    \label{tab:UCI_small_net}
\end{table*}

\newpage
\subsection{Time Consumption}
\label{sec:app:time_consumption}
In the following analysis, we discuss the additional time consumption associated with using a larger subspace dimension. 
We focus solely on the time spent during the training of the subspace construction.
To investigate this, we optimized a plain SSR model (k=0), which serves as our baseline, and trained our $\text{Semi-Subspace}$ models for $k=1,3,7,15$ using the Algorithm \ref{alg:1} stopping after 50 epochs. 
Note that the weights of these four $\text{Semi-Subspace}$ models are controlled using the Bézier curve, as outlined in Equation~\ref{eq:bez}.
We also excluded data preprocessing and model instantiation from the time computation. 
The following time computations were carried out on a \textit{NVIDIA GeForce RTX 3080 Ti} GPU device. 
Additionally, we optimized the time consumption by fully utilizing the GPU capacity through pre-loading data onto its storage.

If we compare the time consumption of the plain SSR model (k=0) with our $\text{Semi-Subspace}$ models we observe an initial offset of around $2.259s-2.219s=0.039s$.
In addition, we see that our $\text{Semi-Subspace}$ model scales linearly with k with a moderate slope.

\begin{figure}[ht]
    \centering
    \includegraphics[width=0.8\linewidth]{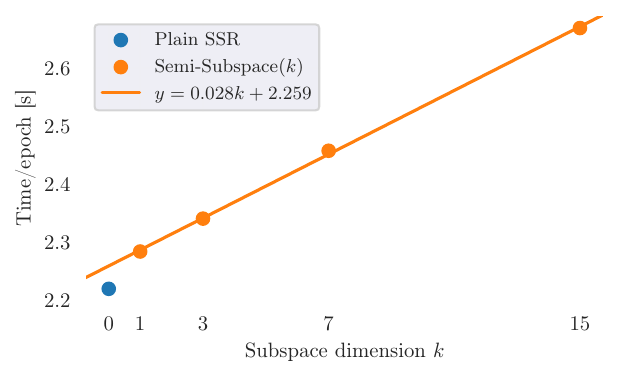} 
    \caption{Time consumption required to optimize the SSR model per epoch depending on the subspace dimension $k$. $k=0$ symbolizes the time consumption to train a plain SSR, whereas $k>0$ depicts the training of the $\text{Semi-Subspace}$ model with respective $k+1$ control points.}
    \label{fig:exp_melanoma_time_consumption}
\end{figure}
\vfill

\end{document}